\documentclass[smallcondensed]{svjour3}     
\smartqed 
\usepackage{graphicx}
\usepackage{subfigure}
\usepackage{booktabs} 

\usepackage{hyperref}

\usepackage{url}
\usepackage{graphicx}  
\usepackage{amsmath}
\usepackage{amssymb}
\usepackage{subfigure}
\usepackage{caption}
\usepackage{algpseudocode}
\usepackage{algorithm}
\usepackage{mathtools} 
\usepackage{enumerate}

\newcommand{\softmax}{\mathrm{softmax}}

\newcommand{\figtop}{{\em (Top)}}
\newcommand{\figbottom}{{\em (Bottom)}}

\newcommand{\algrule}[1][.2pt]{\par\vskip.5\baselineskip\hrule height #1\par\vskip.5\baselineskip}

\begin{document}



\title{Human-Guided Learning of Column Networks:\\ Augmenting Deep Learning with Advice
}

\titlerunning{Human-Guided Learning of Column Networks: Augmenting Deep Learning with Advice}        

\author{Mayukh Das         \and
        Yang Yu   \and    \\
        Devendra Singh Dhami \and \\
        Gautam Kunapuli     \and
        Sriraam Natarajan
}


\institute{M. Das, D.S. Dhami, G. Kunapuli, S. Natarajan  \at
              \email{\{mayukh.das1; devendra.dhami; gautam.kunapuli; sriraam.natarajan\}@utdallas.edu} 
           \and
           Y. Yu \at
           \email{yangyu@hlt.utdallas.edu}
}

\date{Received: date / Accepted: date}

\maketitle

\begin{abstract}
Recently, deep models have been successfully applied in several applications, especially with low-level representations. However, sparse, noisy samples and structured domains (with multiple objects and interactions) are some of the open challenges in most deep models. Column Networks, a deep architecture, can succinctly capture such domain structure and interactions, but may still be prone to sub-optimal learning from sparse and noisy samples. Inspired by the success of human-advice guided learning in AI, especially in data-scarce domains, we propose Knowledge-augmented Column Networks that leverage human advice/knowledge for better learning with noisy/sparse samples. Our experiments demonstrate that our approach leads to either superior overall performance or faster convergence (i.e., both effective and efficient).
\end{abstract}

\section{Introduction}


The re-emergence of Deep Learning~\cite{DeepLearningBook2016} has found significant and successful applications in difficult real-world domains such as image \cite{krizhevsky2012imagenet}, audio \cite{audio} and video processing \cite{videoCVPR}. 
Deep Learning has also been increasingly applied to structured domains, where the data is represented using {\em richer symbolic or graph features} to capture relational structure between entities and attributes in the domain. Intuitively, deep learning architectures are naturally suited to learning and reasoning over such multi-relational domains as they are able to capture increasingly complex interactions between features with deeper layers. However, the combinatorial complexity of reasoning over a large number of relations and objects has remained a significant bottleneck to overcome. 

Recent work in relational deep learning has sought to address this particular issue. This includes relational neural networks \cite{KazemiPoole18-RelNNs,SourekEtAl-15-LRNNs}, relational Restricted Boltzmann machines \cite{KaurEtAl18-RRBM} and neuro-symbolic architectures such as C-ILP \cite{FrancaEtAl-CILP-14}. In our work, we focus upon the framework of {\bf Column Networks} (CLNs) developed by \cite{pham2017column}. Column networks are composed of several (feedforward) mini-columns each of which represents an entity in the domain. Relationships between two entities are modeled through edges between mini-columns. These edges allow for the short-range exchange of information over successive layers of the column network; however, the true power of column networks emerges as the depth of interactions increases, which allows for the natural modeling of long-range interactions. 

Column networks are an attractive approach for several reasons: (1) hidden layers of a CLN share parameters, which means that making the network deeper does not introduce more parameters, (2) as the depth increases, the CLN can begin to model feature interactions of considerable complexity, which is especially attractive for relational learning, and (3) learning and inference are linear in the size of the network and the number of relations, which makes CLNs highly efficient. However, like other deep learning approaches, CLNs rely on vast amounts of data and incorporate little to no knowledge about the problem domain. While this may not be an issue for low-level applications such as image or video processing, it is a significant issue in relational domains, since the relational structure encodes rich, semantic information. This suggests that ignoring domain knowledge can considerably hinder generalization.

It is well known that biasing learners is necessary in order to allow them to inductively leap from training instances to true generalization over new instances ~\cite{Mitchell80}. Indeed, the inductive bias towards ``simplicity and generality'' leads to network architectures with simplifying assumptions through regularization strategies that aim to control the complexity of the neural/deep network. 
While deep learning does incorporate one such bias in the form of  domain knowledge (for example, through parameter tying or convolution, which exploits neighborhood information), we are motivated to develop systems that can incorporate richer and more general forms of domain knowledge. This is especially germane for deep relational models as they inherently construct and reason over richer representations. Such domain-knowledge-based inductive biases have been applied to a diverse array of machine learning approaches, variously known as advice-based, knowledge-based or human-guided machine learning. 

One way in which a human can guide learning is by providing {\em rules over training examples and features}. The earliest such approaches combined explanation-based learning (EBL-NN, \cite{shavlik89ebnn}) or symbolic domain rules with ANNs (KBANN, \cite{towell1994knowledge}). Domain knowledge as {\em rules over input features} can also be incorporated into support vector machines (SVMs, \cite{Cortes1995,Scholkopf98,fung2003knowledge,LeSmolaGartner06,kunapuli2010online}).
Another natural way a human could guide learning is by expressing {\em preferences} and has been studied extensively within the preference-elicitation framework due to Boutilier et al. (\cite{BoutilierEtAl06}). We are inspired by this form of advice as they have been successful within the context of inverse reinforcement learning \cite{KunapuliEtAl13}, imitation learning \cite{odomaaai15} and planning \cite{DasEtAl18}. 

These approaches span diverse machine learning formalisms, and they all exhibit the same remarkable behavior: {\bf better generalization with fewer training examples} because they effectively exploit and incorporate domain knowledge as an inductive bias. This is the prevailing motivation for our approach: to develop a framework that {\bf allows a human to guide deep learning} by incorporating rules and constraints that define the domain and its aspects. Incorporation of prior knowledge into deep learning has begun to receive interest recently, for instance, the recent work on incorporating prior knowledge of color and scene information into deep learning for image classification \cite{DingEtAl18}. 
However, in many such approaches, the guidance is not through a human, but rather through a pre-processing algorithm to generate guidance. Our framework is much more general in that a human provides guidance during learning. Furthermore, the human providing the domain advice is not an AI/ML expert but rather a domain expert who provides rules naturally. We exploit the rich representation power of relational methods to capture, represent and incorporate such rules into relational deep learning models. 


We make the following contributions: (1) we propose the formalism of Knowledge-augmented Column Networks, (2) we present, inspired by previous work (such as KBANN), an approach to inject generalized domain knowledge in a CLN and develop the learning strategy that exploits this knowledge, and (3) we demonstrate, across four real problems in some of which CLNs have been previously employed, the effectiveness and efficiency of injecting domain knowledge. Specifically, our results across the domains clearly show statistically superior performance with small amounts of data. As far as we are aware, this is the first work on human-guided CLNs. 

The rest of the paper is organized as follows. We first review the background necessary for the paper including CLNs. Then we present the formalism of KCLNs and demonstrate with examples how to inject knowledge into CLNs. Next, we present the experimental results across four domains before concluding the paper by outlining areas for future research.

\section{Background and Related Work}
The idea of several processing layers to learn increasingly complex abstractions of the data was initiated by the perceptron model \cite{rosenblatt1958perceptron} and was further strengthened by the advent of the back-propagation algorithm \cite{lecun1998gradient}. A  deep architecture was proposed by \cite{krizhevsky2012imagenet} and  have since been adapted for different problems across the entire spectrum of domains, such as, Atari games via deep reinforcement learning \cite{mnih2013playing}, sentiment classification \cite{glorot2011domain} and image super-resolution \cite{dong2014learning}. 

Applying advice to models has been a long explored problem to construct more robust models to noisy data  \cite{fung2003knowledge,LeSmolaGartner06,towell1994knowledge,kunapuli2010adviceptron,odom2018human}. 
\cite{fu1995introduction} presents a unified view of different variations of knowledge-based neural networks, namely, rule based, decision tree based and semantic constraints based neural networks. Rule-based approaches translate symbolic rules to neural architectures, decision tree based ones impose bounded regions in the parameter space and in constraint  based neural networks, each node denotes a concept and each edge denotes relationships between these concepts.Such advice based learning has been proposed for support vector machines \cite{fung2003knowledge,le2006simpler} in propositional cases and probabilistic logic models \cite{odom2018human} for relational cases. There has also been some work on applying advice to neural networks. \cite{towell1994knowledge} introduce the KBANN algorithm which compiles first order logic rules into a neural network and \cite{kunapuli2010adviceptron} present the first work on applying advice, in the form of constraints, to the perceptron. 
In the rule based neural networks, the data attributes are assigned as input nodes, the target concept(s) as the output nodes and the intermediate concept(s) as the hidden nodes. The decision tree based network inherits its structure from the underlying decision tree. Each decision node in the tree can be viewed as an input space hyperplane and a decision region bounded by these hyperplanes can be viewed as a leaf node. In constraint  based neural network, each node denotes a concept and each edge denotes relationships between these concepts. 
The knowledge-based neural network framework has been applied successfully to various real world problems such as recognizing genes in DNA sequences \cite{noordewier1991training}, microwave design \cite{wang1997knowledge}, robotic control \cite{handelman1990integrating} and recently in personalised learning systems \cite{melesko2018semantic}.
Combining relational (symbolic) and deep learning methods has recently gained significant research thrust since relational approaches are indispensable in faithful and explainable modeling of implicit domain structure, which is a major limitation in most deep architectures in spite of their success. 
While extensive literature exists that aim to combine the two \cite{sutskever2009modelling,rocktaschel2014low,lodhi2013deep,battaglia2016interaction}, to the best of our knowledge, there has been little or no work on incorporating the advice in any such framework.

Column networks transform relational structures into a deep architecture in a principled manner and are designed especially for collective classification tasks~\cite{pham2017column}. 
The architecture and formulation of the column network are suited for adapting it to the advice framework. 
The GraphSAGE algorithm \cite{hamilton2017inductive} shares similarities with column networks since both architectures operate by aggregating neighborhood information but differs in the way the aggregation is performed. Graph convolutional networks \cite{kipf2016semi} is another architecture that is very similar to the way CLN operates, again differing in the aggregation method. \cite{diligenti2017semantic} presents a method of incorporating constraints, as a regularization term, which are first order logic statements with fuzzy semantics, in a neural model and can be extended to collective classification problems. 
While it is similar in spirit to our proposed approach it differs in its representation and problem setup.

{Several recent approaches aim to make deep architectures robust to label noise. They include (1.) learning from easy samples (w/ small loss) by using MentorNets which are neural architectures that estimate curriculum (\textit{i.e.} importance weight on samples)~\cite{jiang2018mentornet}, (2.) noise-robust loss function via additional noise adaptation layers~\cite{goldberger2016training} or via multiplicative modifiers over the error/network parameters~\cite{patrini2017making} and (3) introduction of a regularizer in the loss function for smoothing in presence of adversarial randomizations on the distribution of the response variable~\cite{miyato2018virtual}.}

While the above approaches enable effective learning of deep models in presence of noise, there are some fundamental differences between our problem setting and these related approaches. 
\begin{itemize}
    \item 
    \textbf{{[Type of noise]}}: Our approach aims to handle a specific type of noise, namely systematic/targeted noise \cite{odom2018human}. It occurs frequently in real-world data due to several factors including cognitive bias of humans (or errors in the processes) who record data and sample sparsity.
    \item \textbf{{[Type of error]}}: Systematic noise leads to generalization errors in the learned model (see Example~\ref{ex:metastasis}).
    \item \textbf{{[Structured data]}}: K-CLN works in the context of structured data (entities/relations). Faithful modeling of structure is crucial in most real domains but has the limitation that the data is inherently sparse (most entities are not related to each other \textit{i.e.}, most relations are false in the real world).
    \item \textbf{{[Noise prior]}}: All the noise handling approaches for deep models mentioned earlier explicitly try to model the noise  either via  prior knowledge of noise distribution or by estimating the same with some proposal distribution. While in adversarial regularization \cite{miyato2018virtual} the learned label distribution is used as proposal for generating perturbations, it requires a lot of data for the learner to converge. K-CLN does not explicitly model noise but allows expert knowledge to guide the learner towards better generalization via an inductive bias.
\end{itemize}

\section{Knowledge-augmented Column Networks}
\begin{figure*}
  \begin{minipage}[b]{\textwidth}
    \centering
    \includegraphics[width=\columnwidth]{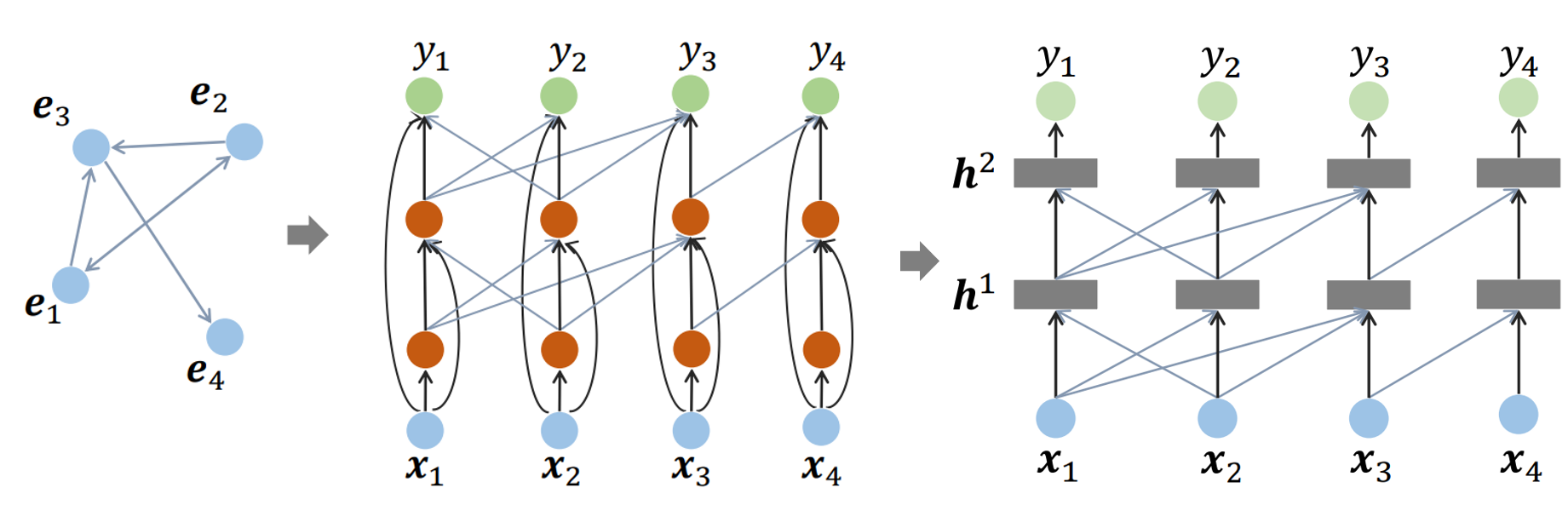}
    \captionof{figure}{Original Column network (diagram source: \cite{pham2017column})}
    \label{fig:CLN}
  \end{minipage}
  \vspace{1.7em}
  \begin{minipage}[b]{\textwidth}
    \centering
    \includegraphics[width=0.8\columnwidth]{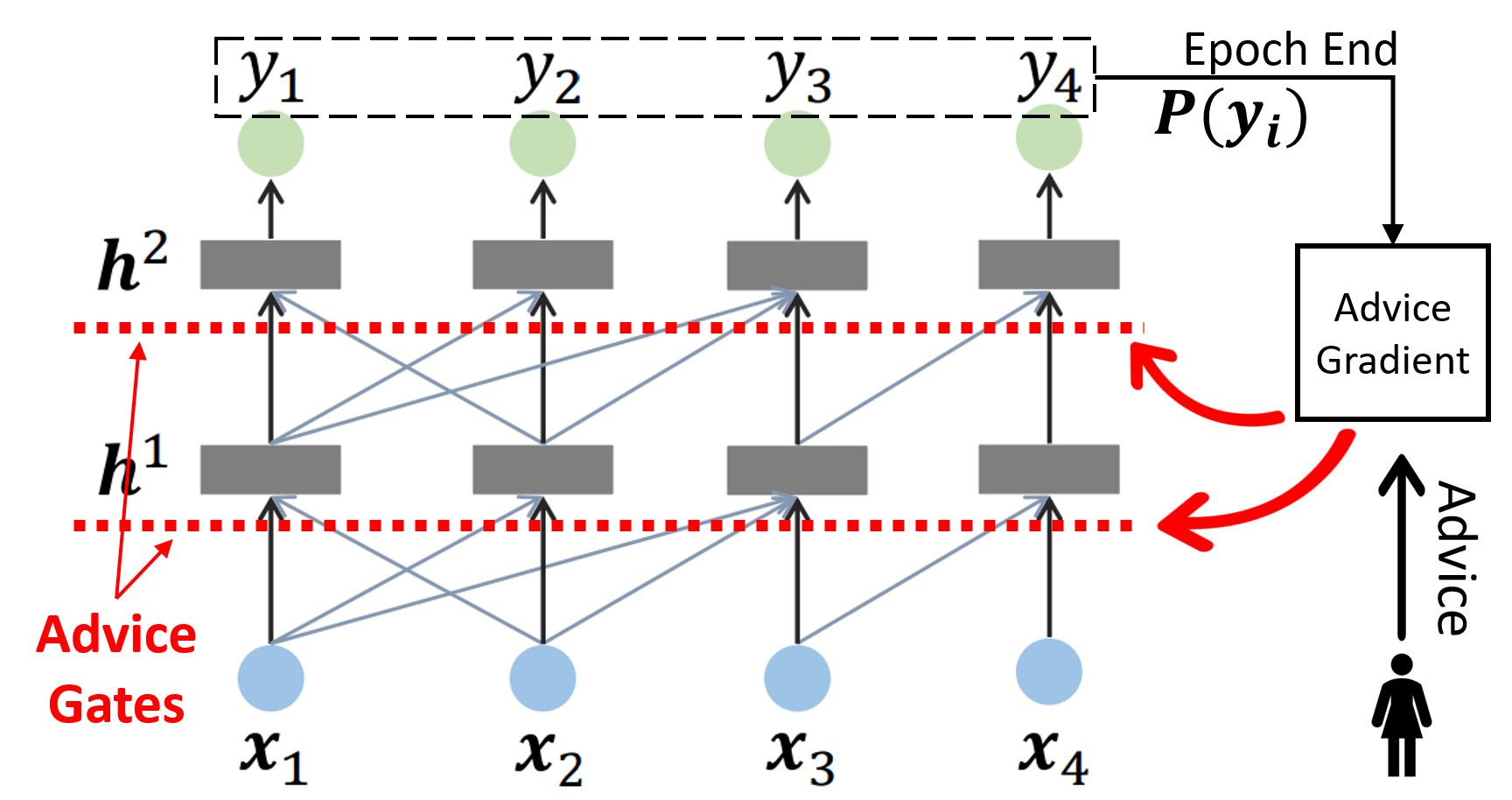}
      \captionof{figure}{Knowledge-augmented Column Network (K-CLN) architecture}
      \label{fig:kcln}
    \end{minipage}
  \end{figure*}

Column Networks~\cite{pham2017column} allow for encoding interactions/relations between entities as well as the attributes of such entities in a principled manner without explicit relational feature construction or vector embedding. This is important when dealing with structured domains, especially, in the case of collective classification. This enables us to seamlessly transform a multi-relational knowledge graph into a deep architecture making them one of the robust \textit{relational} deep models. Figure \ref{fig:CLN} illustrates an example column network, w.r.t. the knowledge graph on the left. Note how each entity forms its own column and relations are captured via the sparse inter-column connectors.

Consider a graph $\mathcal{G}=(V,A)$, where $V = \{e_i\}_{i=1}^{\left|V\right|}$ is the set of vertices/entities. For brevity, we assume only one entity type. However, there is no such theoretical limitation in the formulation. $A$ is the set of arcs/edges between two entities $e_i$ and $e_j$ denoted as $r(e_i,e_j)$. Note that the graph is multi-relational, \textit{i.e.,} $r\in R$ where $R$ is the set of relation types in the domain. To obtain the equivalent Column Network $\mathcal{C}$ from $G$, let $x_i$ be the feature vector representing the attributes of an entity $e_i$ and $y_i$ its label predicted by the model\footnote{Note that since in our formulation every entity is uniquely indexed by $i$, we use $e_i$ and $i$ interchangeably}. $h_i^t$ denotes a hidden node w.r.t. entity $e_i$ at the hidden layer $t$ ($t=1, \ldots, T$  is the index of the hidden layers).
As mentioned earlier, the \textit{context} between 2 consecutive layers captures the dependency of the immediate neighborhood (based on arcs/edges/inter-column connectors). For entity $e_i$, the context w.r.t. $r$ and hidden nodes are computed as, 
\begin{align}
    \label{eqn:hiddencontext}
    &  c_{ir}^t = \frac{1}{\left|\mathcal{N}_r(i)\right|}\sum_{j\in \mathcal{N}_r(i)} h_j^{t-1} \\
    & h_i^t = g\left(b^t + W^t h_i^{t-1} + \frac{1}{z} \sum_{r\in R} V_r^t c_{ir}^t\right)
\end{align}
where $\mathcal{N}_r(i)$ are all the neighbors of $e_i$ w.r.t. $r$ in the knowledge graph $\mathcal{G}$. Note the absence of context connectors between $h^t_2$ and $h^t_4$ (Figure \ref{fig:CLN}, \textit{right}) since there does not exist any relation between $e_2$ and $e_4$ (Figure \ref{fig:CLN}, \textit{left}). 
The activation of the hidden nodes is computed as the sum of the bias, the weighted output of the previous hidden layer and the weighted contexts 
where $W^t \in \mathbb{R}^{K^t \times K^{t−1}}$ and $V^t_r \in R^{K^t\times K^{t−1}}$ are weight parameters and $b^t$ is a bias for some activation function $g$. $z$ is a pre-defined constant that controls the parameterized contexts from growing too large for complex relations. Setting $z$ to the average number of neighbors of an entity is a reasonable assumption.  
The final output layer is a softmax over the last hidden layer. 
\begin{equation}
    \label{eq:op} P(y_i = \ell|h_i^T) = softmax\left( b_l + W_l h_i^T \right)
\end{equation}
where $\ell\in L$ is the label ($L$ is the set of labels) and $T$ is the index of the last hidden layer. 

Following \cite{pham2017column}, we choose to formulate our approach in the context of a relation-sensitive predictive modeling, specifically collective classification tasks. However, structured data is implicitly sparse since most entities in the world are not related to each other, thereby adding to the existing challenge of faithful modeling of the underlying structure. The challenge is amplified as
{\it we aim to learn in the presence of knowledge-rich, data-scarce} problems.
As we show empirically, sparse samples (or targeted noise) may lead to sub-optimal learning or slower convergence. 
\begin{example}
\label{ex:metastasis}
Consider a problem of classifying whether a published article is about carcinoid metastasis \cite{zuetenhorst2005metastatic} or is irrelevant, from a citation network, and textual features extracted from the articles themselves. There are several challenges: (1) Data is implicitly sparse due to rarity of studied cases and experimental findings, (2) Some articles may cite other articles related to carcinoid metastasis and contain a subset of the textual features, but address another topic and (3) Finally, the presence of targeted noise, where some important citations were not extracted properly by some citation parser and/or the abstracts are not informative enough.
\end{example}
The above cases may lead to the model not being able to effectively capture certain dependencies, or converge slower, even if they are captured somewhere in the advanced layers of the deep network. Our approach attempts to alleviate this problem via augmented learning of Column Networks using human advice/knowledge. We formally define our problem in the following manner,

\noindent \fbox{
\parbox{0.97\columnwidth}{
\noindent {\bf Given}: A sparse multi-relational graph $\mathcal{G}$, attributes $x_i$ of each entity (sparse or noisy) in $\mathcal{G}$, equivalent Column-Network $\mathcal{C}$ and access to a Human-expert\\
\noindent{\bf To Do:} More effective and efficient collective classification by knowledge augmented training of $\mathcal{C}(\theta)$, where $\theta = \langle\{W^t\}_1^T, \{V_r^t\}_{r\in R; t=1}^{t=T}, \{W_{\ell}\}_{\ell\in L}\rangle$ is the set of all the network parameters of the Column Network.
}}

We develop \textbf{\textit{K}}nowledge-augmented \textbf{\textit{C}}o\textbf{\textit{L}}umn \textbf{\textit{N}}etworks (K-CLN), that incorporates human-knowledge, for more effective and efficient learning from relational data (Figure~\ref{fig:kcln} illustrates the overall architecture). While knowledge-based connectionist models are not entirely new, our formulation provides - (1) a principled approach for incorporating advice specified in an intuitive logic-based encoding/language (2) a deep model for collective classification in relational data. 
\vspace{-0.5em}
\subsection{Knowledge Representation}
Any model specific encoding of domain knowledge, such as numeric constraints or modified loss functions etc., has several limitations, namely (1) counter-intuitive to the humans since they are domain experts and not experts in machine learning (2) the resulting framework is brittle and not generalizable.  Consequently, we employ preference rules (akin to IF-THEN statements) to capture human knowledge.
\begin{definition}
\label{def:pref}
A preference is a modified Horn clause,
\begin{align}
    \nonumber \mathtt{\land_{k, x} Attr_k(E_x) \land \ldots \land_{r\in R, x, y} r(E_x,E_y)} \Rightarrow \mathtt{[} \mathtt{label(E_z,\ell_1)} \uparrow; \mathtt{label(E_k,\ell_2) \downarrow]}
\end{align}
where $\ell_1,\ell_2 \in L$ and the $\mathtt{E_x}$ are variables over entities, $\mathtt{Attr_k(E_x)}$ are attributes of $E_x$ and $\mathtt{r}$ is a relation. $\mathtt{\uparrow}$ and $\mathtt{\downarrow}$ indicate the preferred non-preferred labels respectively. Quantification is implicitly $\forall$ and hence dropped. We denote a set of preference rules as $\mathfrak{P}$.
\end{definition}
Note that we can always, either have just the preferred label in head of the clause and assume all others as non-preferred, or assume the entire expression as a single literal. Intuitively a rule can be interpreted as conditional rule, \textbf{IF [conditions hold] THEN label $\mathtt{\ell}$ is preferred}. A preference rule can be partially instantiated as well, \textit{i.e.}, or more of the variables may be substituted with constants. 

\begin{example}
For the prediction task mentioned in Example~\ref{ex:metastasis}, a possible preference rule could be,
\begin{align}
    \nonumber \mathtt{hasWord(E_1, ``AI")} \land \mathtt{hasWord(E_2,``domain")} \land \mathtt{cites(E_2, E_1)}\\
    \nonumber \Rightarrow \mathtt{label(E_2,``irrelevant")\uparrow}
\end{align}
Intuitively, this rule denotes that an article is not a relevant clinical work to carcinoid metastasis if it cites an `AI' article and contains the word ``domain", since it is likely to be another AI article that uses carcinoid metastatis as an evaluation domain.
\end{example}

\subsection{Knowledge Injection}
Given that knowledge is provided as \textit{partially-instantiated} preference rules $\mathfrak{P}$, more than one entity may satisfy a preference rule. Also, more than one preference rules may be applicable for a single entity.
The main intuition is that we aim to consider the error of the trained model w.r.t. both the data and the advice. Consequently, in addition to the \textit{``data gradient"} as with original CLNs, there is a \textit{``advice gradient''}. This gradient acts a feedback to augment the learned weight parameters (both column and context weights) towards the direction of the \textit{advice gradient}. It must be mentioned that not all parameters will be augmented. Only the parameters w.r.t. the entities and relations (contexts) that satisfy $\mathfrak{P}$ should be affected. 
Let $\mathcal{P}$ be the set of entities and relations that satisfy the set of preference rules $\mathfrak{P}$. The hidden nodes (equation~\ref{eqn:hiddencontext}) can now be expressed as,
\begin{align}
  \label{eq:modhidden} \nonumber   h_i^t = g\left(b^t + W^t h_i^{t-1} \Gamma^{(W)}_i + \frac{1}{z} \sum_{r\in R} V_r^t c_{ir}^t \Gamma^{(c)}_{ir}\right)\\
  \text{s.t.}~ \Gamma_i, \Gamma_{i,r} = 
 \begin{cases}
                                   1 & \text{if $i,r \notin \mathcal{P}$} \\
                                \mathcal{F}(\alpha\nabla_i^{\mathfrak{P}}) & \text{if $i,r \in \mathcal{P}$}
  \end{cases}
\end{align}
where $i \in \mathcal{P}$ and $\Gamma^{(W)}_i$ and $\Gamma^{(c)}_{ir}$ are advice-based soft gates with respect to a hidden node and its context respectively. $\mathcal{F}()$ is some gating function, $\nabla_i^{\mathfrak{P}}$ is the \textit{``advice gradient''} and $\alpha$ is the trade-off parameter explained later. The key aspect of soft gates is that they attempt to enhance or decrease the contribution of particular edges in the column network aligned with the direction of the \textit{``advice gradient''}. We choose the gating function $\mathcal{F}()$ as an exponential $[\mathcal{F}(\alpha\nabla_i^{\mathfrak{P}}) = \exp{(\alpha\nabla_i^{\mathfrak{P}})}]$. The intuition is that soft gates are natural, as they are multiplicative and a positive gradient will result in $\exp{(\alpha\nabla_i^{\mathfrak{P}})} > 1$ increasing the value/contribution of the respective term, while a negative gradient results in $\exp{(\alpha\nabla_i^{\mathfrak{P}})} < 1$ pushing them down. We now present the \textit{``advice gradient''} (the gradient with respect to preferred labels). 

\begin{proposition}
\label{eq:grad}
Under the assumption that the loss function with respect to advice / preferred labels is a log-likelihood, of the form $\mathcal{L^\mathfrak{P}} = \log P(y_i^{(\mathfrak{P})}|h_i^T)$, then the advice gradient is,
$ \nabla_i^{\mathfrak{P}} = I({y_i^{(\mathfrak{P})}}) - P(y_i)$, 
where $y_i^{(\mathfrak{P})}$ is the preferred label of entity and $i\in \mathcal{P}$ and $I$ is an indicator function over the preferred label. For binary classification, the indicator is inconsequential but for multi-class scenarios it is essential ($I = 1$ for preferred label $\ell$ and $I=0$ for $L\setminus \ell$).
\end{proposition}
 Since an entity can satisfy multiple advice rules we take the \textit{MAX} preferred label, \textit{i.e.}, we take the label $y_i^{(\mathcal{P})}=\ell$ to the preferred label if $\ell$ is given by most of the advice rules that $e_j$ satisfies. In case of conflicting advice (i.e. different labels are equally advised), we simply set the advice label to be the label given by the data, $y_i^{(\mathfrak{P})}=y_i$. 

\noindent\textit{\textbf{Proof Sketch:}} Most advice based learning methods formulate the effect of advice as a constraint on the parameters or a regularization term on the loss function. We consider a regularization term based on the advice loss $\mathcal{L}^{(\mathfrak{P})} = \log P(y_i=y_i^{(\mathfrak{P})}|h_i^T)$ and we know that $P(y_i|h_i^T) = \softmax(b_\ell + W_\ell h_i^T)$. We consider $b_\ell + W_\ell h_i^T = \Psi_{(y_i,h_i^T)}$ in its functional form following prior non-parametric boosting approaches~\cite{odomaaai15}. Thus $P(y_i=y_i^{(\mathfrak{P})}|h_i^T) = \exp{(\Psi_{(y_i^{(\mathfrak{P})},h_i^T)})}/\sum_{y' \in {L}} \exp{(\Psi_{(y',h_i^T)})}$. A functional gradient w.r.t. $\Psi$ of $\mathcal{L}^{(\mathfrak{P})}$ yields, 
$$\nabla_i^{\mathfrak{P}} = \frac{\partial \log P(y_i=y_i^{(\mathfrak{P})}|h_i^T)}{\partial \Psi_{(y_i^{(\mathfrak{P})},h_i^T)}}  = I({y_i^{(\mathfrak{P})}}) - P(y_i)$$
Alternatively, assuming a squared loss such as $( y_i^{(\mathfrak{P})} - P(y_i))^2$, would result in an advice gradient of the form $2( y_i^{(\mathfrak{P})} - P(y_i))(1-P(y_i))P(y_i)$.

As illustrated in the K-CLN architecture (Figure~\ref{fig:kcln}), at the end of every epoch of training the \textit{advice gradients} are computed and soft gates are used to augment the value of the hidden units as shown in Equation~\ref{eq:modhidden}. 

\begin{proposition}
\label{prop:balance}
Given that the loss function $\mathcal{H}_i$ of original CLN is cross-entropy (binary or sparse-categorical for the binary and multi-class prediction cases respectively) and the objective \textit{w.r.t.} advice is log-likelihood, the functional gradient of the modified objective for K-CLN is, 
\begin{align}
   \nonumber \nabla(\mathcal{H}'_i) & = (1-\alpha)\left(y_iI - P(y_i|h^T)\right) + \alpha \left(I_i^{\mathfrak{P}}-P(y_i^{\mathfrak{P}}|h^T)\right)\\
   \label{eq:modgrad} & = (1-\alpha)\nabla_i + \alpha \nabla_i^{\mathfrak{P}}
\end{align}
where $0\leq\alpha\leq 1$ is the trade-off parameter between the effect of data and effect of advice, $I_i$ and $I_i^{\mathfrak{P}}$ are the indicator functions on the label w.r.t. the data and the advice respectively and $\nabla_i$ and $\nabla_i^{\mathfrak{P}}$ are the gradients, similarly, w.r.t. data and advice respectively.
\end{proposition}

\noindent \textit{\textbf{Proof Sketch:}} The original objective function (\textit{w.r.t.} data) of CLNs is cross-entropy. For clarity, let us consider the binary prediction case, where the objective function is now a binary cross-entropy of the form\\ $\mathcal{H} = -\frac{1}{N}\sum_{i=1}^N y_i\log(P(y_i)) + (1-y_i)\log(1-P(y_i))$.

Ignoring the summation for brevity, for every entity $i$, $\mathcal{H}_i = y_i\log(P(y_i)) + (1-y_i)log(1-P(y_i))$. Extension to the multi-label prediction case with a sparse categorical cross-entropy is straightforward and is an algebraic manipulation task. Now, from Proposition~\ref{eq:grad}, the loss function \textit{w.r.t.} advice is the log likelihood of the form, $\mathcal{L}^{\mathfrak{P}} = \log P(y_i^{\mathfrak{P}}|h^T)$. Thus the modified objective is expressed as,
\begin{align}
    \label{eqn:modobj} \mathcal{H}_i' = (1-\alpha) \left[ y_i\log\left(P(y_i)\right) + (1-y_i)log\left(1-P(y_i)\right)\right] + \alpha \log(P(y_i^{\mathfrak{P}}))
\end{align}
where $\alpha$ is the trade-off parameter. $P(y) = P(y|h^T)$ can be implicitly understood. Now we know from Proposition~\ref{eq:grad} that the distributions, $P(y_i)$ and $P(y_i^\mathfrak{P})$, can be expressed in their functional forms, given that the activation function of the output layer is a \textit{softmax}, as $P(y_i)= \exp{(\Psi_{(y_i,h_i^T)})}/\sum_{y' \in {L}} \exp{(\Psi_{(y',h_i^T)})}$. Taking the functional (partial) gradients (\textit{w.r.t.} $\Psi_{(y_i,h_i^T)}$ and $\Psi_{(y_i^{\mathfrak{P}},h_i^T)}$) of the modified objective function (Equation~\ref{eqn:modobj}), followed by some algebraic manipulation we get,
\begin{align}
    \nonumber \nabla(\mathcal{H}'_i) & = (1-\alpha) [y_iI_i - y_i P(y_i) - P(x_i) + y_i P(y_i)] + \alpha (I_i^{\mathfrak{P}}-P(y_i^{\mathfrak{P}}))\\
    \nonumber & = (1-\alpha)\left(y_iI - P(y_i)\right) + \alpha \left(I_i^{\mathfrak{P}}-P(y_i^{\mathfrak{P}})\right) &\text{(Eqn~\ref{eq:modgrad})}
\end{align}

Hence, it follows from Proposition~\ref{prop:balance} that the data and the advice balances the training of the K-CLN network parameters $\theta^\mathfrak{P}$ via the trade-off hyperparameter $\alpha$. When data is noisy (or sparse with negligible examples for a region of the parameter space) the advice (if correct) induces a bias on the output distribution towards the correct label. Even if the advice is incorrect, the network still tries to learn the correct distribution to some extent from the data (if not noisy). The contribution of the effect of data versus the effect of advice will primarily depend on $\alpha$. If both the data and human advice are sub-optimal (noisy), the correct label distribution is not even learnable.

\begin{algorithm}[t]
\begin{algorithmic}[1]
\Procedure{KCLN}{Knowledge graph $\mathcal{G}$, Column network $\mathcal{C}(\theta)$, Advice $\mathfrak{P}$, Trade-off $\alpha$}
\State K-CLN $\mathcal{C}^{\mathfrak{P}}(\theta^{\mathfrak{P}}) \gets \mathcal{C}(\theta)$ \Comment{w/ changed expr. of hidden units w.r.t. Eqn~\ref{eq:modhidden}}
\State Initialize $\theta^{\mathfrak{P}} \gets \{0\}$ \Comment{n/w parameters of K-CLN intialized to $0$}
\State $\mathcal{M}^\mathcal{P} = \langle\mathcal{M}^W,\mathcal{M}^c,\mathcal{M}^{label}\rangle \gets$ \Call{CreateMask}{$\mathcal{G},\mathfrak{P}$} \Comment{mask $\forall$ ents./rels. $\in \mathcal{P}$}
\State Initial gradients $\forall i ~ \mathbf{\nabla}_{i,0}^{\mathfrak{P}} = 0$; $i \in \mathcal{P}$ \Comment{advice gradient $=0$ at epoch $=0$}
\For{epochs k=1 to convergence} \Comment{convergence criteria same as original CLN}
\State Get advice gradients $\nabla_{i,(k-1)}^{\mathfrak{P}}$ w.r.t. previous epoch $k-1$
\State Gates $\Gamma^{\mathfrak{P}}_i, \Gamma^{\mathfrak{P}}_{i,r} \gets \exp{(\alpha \nabla_i^{\mathfrak{P}}\times \mathcal{M}_i^\mathcal{P})}$  \Comment{$\mathcal{M}^W$ and $\mathcal{M}^c$}
\State Train $\mathcal{C}^{\mathfrak{P}}$ using Equation~\ref{eq:modhidden}; Update $\theta^{\mathfrak{P}}$
\State Compute $\forall i ~ P(y_i)$ from $\mathcal{C}^{\mathfrak{P}}$ \Comment{for current epoch $k$}
\State Store $\forall i ~ \nabla_{i,k}^{\mathfrak{P}} \gets I({y_i^{(\mathfrak{P})}}) - P(y_i)$  \Comment{get $ I({y_i^{(\mathfrak{P})}})$ from $\mathcal{M}^{label}$}
\EndFor\\
\Return {K-CLN $C^{\mathfrak{P}}$}
\EndProcedure
\algrule
\Procedure{CreateMask}{Knowledge graph $\mathcal{G}$,Advice $\mathfrak{P}$}
\State $\mathcal{M}^{W}[D\times \left|O\right|] \gets \emptyset$ \Comment{$D$:  feature length of entity; $\left|O\right|$: \# entities where $\mathcal{G}=(O,R)$}
\State $\mathcal{M}^{c}[\left|O\right|\times \left|O\right|] \gets \emptyset$
\State $\mathcal{M}^{label}[ \left|O\right|\times L] \gets \emptyset$ \Comment{where $L$ is the number of distinct labels}
\Statex \Comment{$\mathcal{M}^{W}$: entity mask; $\mathcal{M}^{c}$: context mask \& $\mathcal{M}^{label}$: label mask, w.r.t. advice}
\For{each preference $p \in \mathfrak{P}$}
\If{$\forall i \in O \land \forall r \in R$ : $i$ and $r$ satisfies $p$}
\State $\mathcal{M}^W[x,i] \gets 1$ \Comment{where $x$ is the feature affected by $p$}
\State $\mathcal{M}^c[i,j] \gets 1$ \Comment{where $r = \langle i,j\rangle \in R; j\neq i; j \in O $} 
\State $\mathcal{M}^{label}[i,\ell] \gets 1$; where \Call{LabelOf}{$i|p$} $= \ell$ 
\EndIf
\EndFor
\State \Return{$\langle\mathcal{M}^W,\mathcal{M}^c,\mathcal{M}^{label}\rangle$}
\EndProcedure
\end{algorithmic}
\caption{\textsc{K-CLN}: \underline{\textbf{K}}nowledge-augmented \underline{\textbf{C}}o\underline{\textbf{L}}umn \underline{\textbf{N}}etworks}
\label{algo:kcln}
\end{algorithm}

\subsection{The Algorithm}
Algorithm~\ref{algo:kcln} outlines the key steps involved in our approach. \texttt{KCLN()} is the main procedure [\textbf{lines: 1-14}] that trains a Column Network using both the data (the knowledge graph $\mathcal{G}$) and the human advice (set of preference rules $\mathfrak{P}$). It returns a K-CLN $\mathcal{C}^{\mathfrak{P}}$ where $\theta^{\mathfrak{P}}$ are the network parameters, which are initialized to any arbitrary value ($0$ in our case; [\textbf{line: 3}]). As described earlier, the network parameters of K-CLN (same as CLN) are manipulated (stored and updated) via tensor algebra with appropriate indexing for entities and relations. Also recall that our gating functions are piece-wise/non-smooth and apply only to the subspace of entities, features and relations where the preference rules are satisfied. Thus, as a pre-processing step, we create tensor masks that compactly encode such a subspace with a call to the procedure \textsc{CreateMask()} [\textbf{line: 4}], which we elaborate later. 

The network $\mathcal{C}^{\mathfrak{P}}(\theta^{\mathfrak{P}})$ is then trained through multiple epochs till convergence [\textbf{lines: 6-12}]. At the end of every epoch the output probabilities and the gradients are computed and stored in a shared data structure [\textbf{line: 11}] such that they can be accessed subsequently to compute the advice gates [\textbf{lines: 7-8}]. Our network is trained largely similar to the original CLN with \emph{two key modifications} [\textbf{line: 9}], namely,
\begin{enumerate}
    \item Equation~\ref{eq:modhidden} is used as the modified expression for hidden units
    \item The data trade-off $1-\alpha$ is multiplied with the original loss while its counterpart, the advice trade-off $\alpha$, is used to compute the gates [\textbf{line: 8}]
\end{enumerate}

Procedure \textsc{CreateMask()} [\textbf{lines: 15-27}] constructs the tensor mask(s) over the space of entities, features and relations/contexts that are required to compute the gates (as seen in \textbf{line: 8}). Data (the ground knowledge graph $\mathcal{G}$) and the set of preference rules $\mathfrak{P}$ are provided as inputs. There are {\em three key components} of the advice mask. They are, 
\begin{enumerate}
    \item The entity mask $\mathcal{M}^W$ (a tensor of dimensions - \#entities by length of feature vector)  that indicates which entities and the relevant features are affected by the advice/preference
    \item The context mask $\mathcal{M}^c$ (\#entities by \#entities) which indicates the contexts that are affected (relations are directed and so this matrix is asymmetric)
    \item The label mask $\mathcal{M}^{label}$ which indicate the preferred label of the affected entities, in one-hot encoding
\end{enumerate}
All the components are initialized to zeros. 
The masks are then computed for every preference rule iteratively [\textbf{lines: 19-25}]. This includes satisfiability checking for a given preference rule $p\in\mathfrak{P}$ [\textbf{line: 20}], which is achieved via subgraph matching on the knowledge graph $\mathcal{G}$ since a preference rule (Horn clause - Definition~\ref{def:pref}) can be viewed a subgraph template. For more details, we refer to the work on employing hyperraph/graph databases for counting instances of horn clauses~\cite{richards95,dasSDM2016,DasAAAI19}. So for all entities, relevant features, and relations/contexts that satisfy the rule, the corresponding elements of the tensor masks are set to $1$ [\textbf{lines: 21-23}]. The components $\mathcal{M}^W$ and $\mathcal{M}^c$ are used in gate computation in the KCLN procedure and $\mathcal{M}^{label}$ is used for the indicator $I_i^\mathfrak{P}$ in the advice gradient.

After considering the formulation and the learning of KCLNs, we now turn our attention to empirical evaluation of the proposed work. 

\section{Experiments}
We investigate the following questions as part of our experiments, - 
\begin{enumerate}
    \item[\textbf{Q1}] Can K-CLNs learn effectively with noisy sparse samples i.e., performance?
    \item[\textbf{Q2}] Can K-CLNs learn efficiently with noisy sparse samples i.e., speed of learning?
    \item[\textbf{Q3}] How does quality of advice affect the performance of K-CLN i.e., reliance on robust advice?
\end{enumerate}
We compare against the original Column Networks architecture with no advice\footnote{Vanilla CLN indicates the original Column Network architecture \cite{pham2017column}} as a baseline. Our intention is to show how advice/knowledge can guide model learning towards better predictive performance and efficiency, in the context of collective classification using Column Networks. Also, we have discussed earlier, in detail, how our problem setting is distinct from most existing noise robust deep learning approaches. Hence, we restricted our comparisons to the original work.

\subsection{Experimental Setup}
\noindent \textbf{System:}
K-CLN has been developed by extending original CLN architecture, which uses \textit{Keras} as the functional deep learning API with a \textit{Theano} backend for tensor manipulation. We extend this system to include: (1) advice gradient feedback at the end of every epoch, (2) modified hidden layer computations and (3) a pre-processing wrapper to parse the advice/preference rules and create appropriate tensor masks. Since it is not straightforward to access final layer output probabilities from inside any hidden layer using keras, we use \textit{Callbacks} to write/update the predicted probabilities to a shared data structure at the end of every \textit{epoch}. This data structure is then fed via inputs to the hidden layers. Each mini-column with respect to an entity is a dense network of $10$ hidden layers with $40$ hidden nodes  in each layer (similar to the most effective settings outlined in \cite{pham2017column}). 

The pre-processing wrapper acts as an interface between the advice encoded in a symbolic language (horn clauses) and the tensor-based computation architecture. The \textit{advice masks} encode $\mathcal{P}$, \textit{i.e.}, the set of entities and contexts where the gates are applicable (Algorithm~\ref{algo:kcln}).

\noindent\textbf{Domains:} We evaluate our approach on {\bf four relational} domains -- \textit{Pubmed Diabetes} and \textit{Corporate Messages}, which are multi-class classification problems, and \textit{Internet Social Debates} and \textit{Social Network Disaster Relevance}, which are binary. \textit{Pubmed Diabetes}\footnote{\url{https://linqs.soe.ucsc.edu/data}} is a citation network for predicting whether a peer-reviewed article is about \textit{Diabetes Type 1, Type 2 or none}, using textual features (TF-IDF vectors) from $19717$ pubmed abstracts as well as $44,338$ citation relationships between them. It comprises  articles, considered as an entities, with $500$ bag-of-words textual features (TF-IDF weighted word vectors), and $44,338$ citation relationships among each other. 
\textit{Internet Social Debates}\footnote{\url{http://nldslab.soe.ucsc.edu/iac/v2/}} is a data set for predicting stance (`for'/`against') about a debate topic from online posts on social debates. It contains $6662$ posts (entities) characterized by TF-IDF vectors, extracted from the text and header, and $\sim 25000$ relations of 2 types, {`sameAuthor'} and {`sameThread'}. \textit{Corporate Messages}\footnote{\url{https://www.figure-eight.com/data-for-everyone/}} is an intention prediction data set of $3119$ flier messages sent by corporate groups in the finance domain with $1,000,000$ \textit{sameSourceGroup} relations. The target is to predict the intention of the message \textit{(Information, Action or Dialogue)}. 
Finally, \textit{Social Network Disaster Relevance} (same source) is a relevance prediction data set of 8000 \textit{Twitter} posts, curated and annotated by crowd with their relevance scores. Along with bag-of-word features we use confidence score features and $35k$ relations among tweets (of types \textit{`same author'} and \textit{`same location'}). Table~\ref{tab:doms} outlines the important aspects of the 4 domains (data sets) used in our experimental evaluation. As indicated earlier, inspired by original CLNs, we evaluate our approach on both binary and multi-class prediction problems.

\begin{table}[t]
    \centering
    \begin{tabular}{|l|c|c|c|c|}
    \hline
       \textbf{Domain/Data set}  &  \textbf{\#Entities} & \textbf{\#Relations} & \textbf{\#Features} & \textbf{Target type}\\
       \hline
       \hline
        Pubmed Diabetes & $19717$ & $44,338$ & $500$ & Multi-class\\
        Corporate Messages & $3119$ & $\sim 1,000,000$ & $750$ & Multi-class\\
        Online Social Debates & $6662$ & $\sim 25000$ & $500$ & Binary\\
        Disaster Relevance & $8000$ & $35000$ & $504$ & Binary\\
        \hline
    \end{tabular}
    \caption{Evaluation domains and their properties}
    \label{tab:doms}
\end{table}

\noindent\textbf{Metrics:} Following \cite{pham2017column}, we report macro-F1 and micro-F1 scores for the multi-class problems,  and F1 scores and AUC-PR for the binary ones. Macro-F1 computes the F1 score independently for each class and takes the average whereas a micro-F1 aggregates the contributions of all classes to compute the average F1 score. For all experiments we use $10$ hidden layers and $40$ hidden units per column in each layer. All results are averaged over 5 runs. Other settings are consistent with original CLN.

\noindent\textbf{Human Advice:} K-CLN is designed to handle arbitrarily complex expert advice given that they are encoded as preference rules. However, even with some relatively simple preference rules K-CLN is more effective in sparse samples. \textit{For instance, in Pubmed, the longest one among the 4 preference rules used is, $\mathtt{HasWord(e_1,`fat')}$ $\land$ $\mathtt{HasWord(e_1,`obese')}$ $\land$ $\mathtt{Cites(e_2,e_1)}$ $\Rightarrow$ $\mathtt{label(e_2, type_2)}\uparrow$}. Note how a simple rule, indicating an article citing another one discussing obesity is likely to be about Type2 diabetes, proved to be effective. Expert knowledge from real physicians can thus, prove to be even more effective.  In \textit{Disaster Relevance} we used rules that did not require much domain expertise, such as \textit{if a tweet is by the same user who usually posts non-disaster tweets then the tweet is likely to be a non-disaster one}. Sub-optimal advice may lead to a wrong direction of the \textit{Advice Gradient}. However, our soft gates do not alter the loss, but instead promote/demote the contribution of nodes/contexts. Similar to Patrini et al.,~\cite{patrini2017making}, data is still balancing the effect of advice during training.

\begin{figure*}
\begin{minipage}{\textwidth}
    \centering
    \subfigure[Micro-F1 (w/ epochs)]{
    \includegraphics[width=0.45\textwidth]{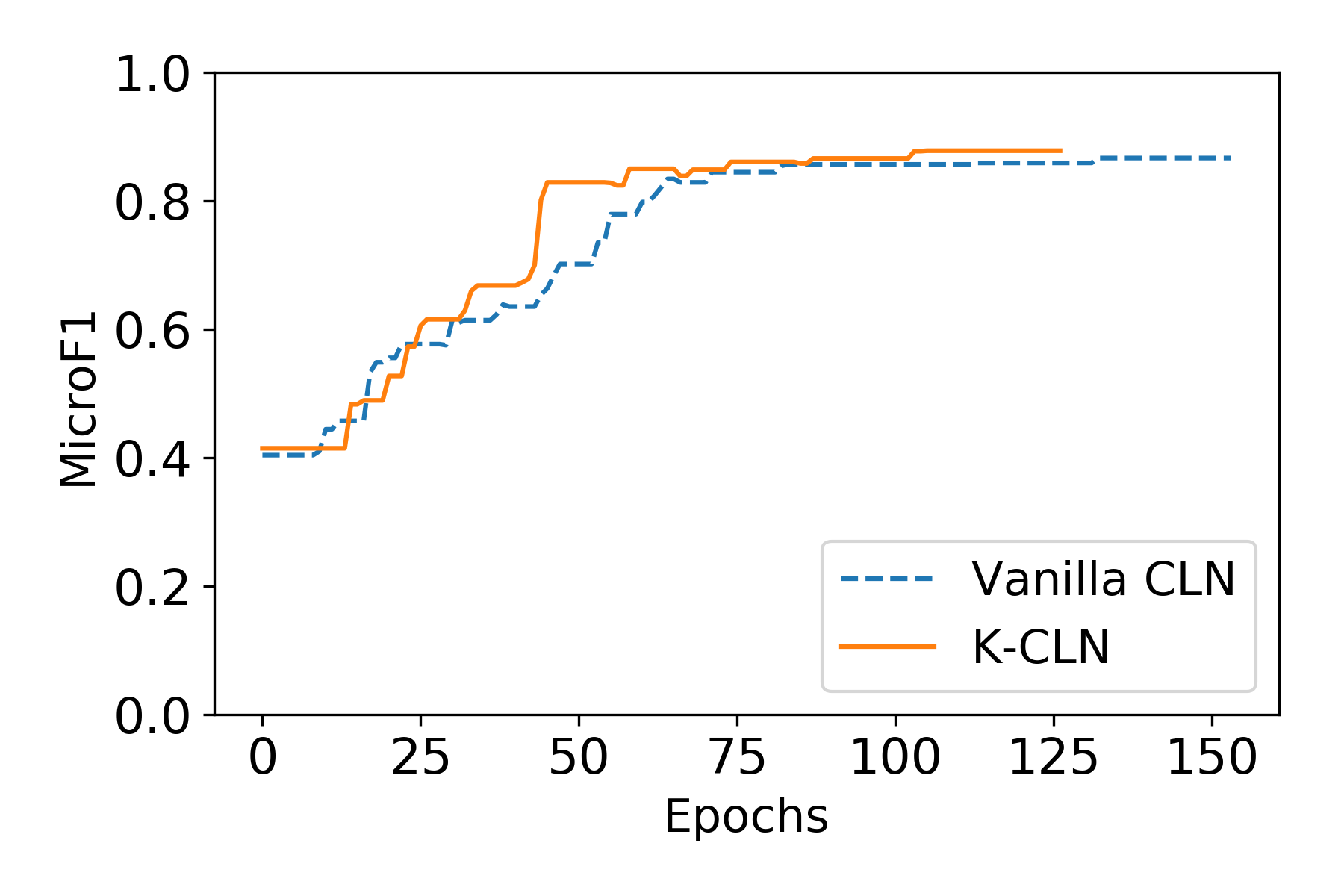}
    \label{fig:microPub}
    }
    \subfigure[Macro-F1 (w/ epochs)]{
    \includegraphics[width=0.45\textwidth]{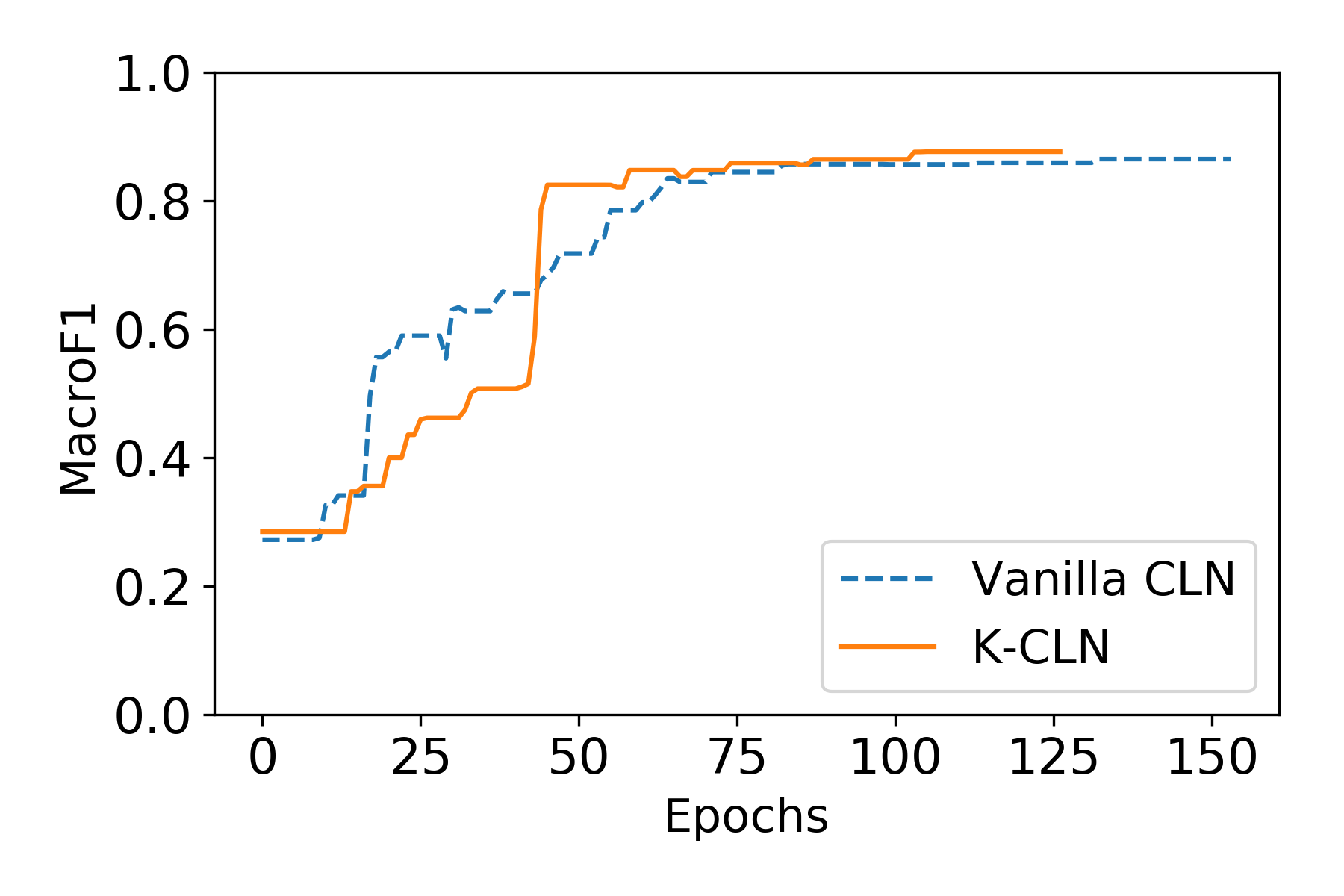}
    \label{fig:macroPub}}\\

    \subfigure[Micro-F1 (w/ varying sample size)]{
    \includegraphics[width=0.45\textwidth]{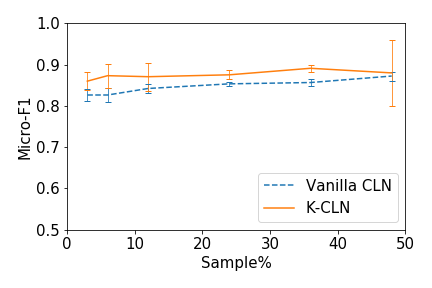}
    \label{fig:microPubSam}
    }
    \subfigure[Macro-F1 (w/ varying sample size)]{
    \includegraphics[width=0.45\textwidth]{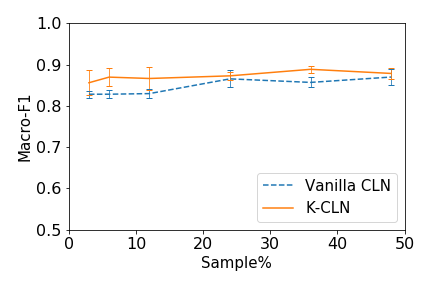}
    \label{fig:macroPubSam}}
    \caption{\textbf{[Pubmed Diabetes publication prediction (multi-class)]} Learning curves  - \figtop ~ w.r.t. training epochs at 24\% (of total) sample, \figbottom ~ w.r.t. varying sample sizes [best viewed in color].}
    \label{fig:PubMed}
\end{minipage}

\begin{minipage}{\textwidth}
    \centering
    \subfigure[Micro-F1 (w/ epochs)]{
    \includegraphics[width=0.45\textwidth]{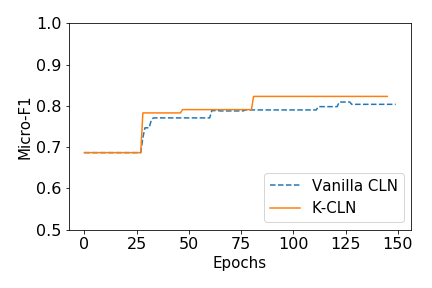}
    \label{fig:microCorp}
    }
    \subfigure[Macro-F1 (w/ epochs)]{
    \includegraphics[width=0.45\textwidth]{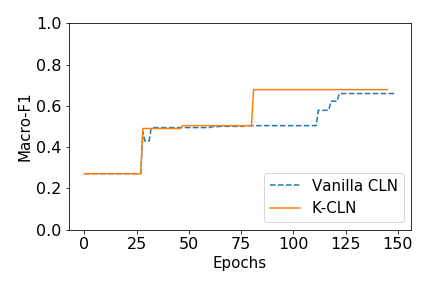}
    \label{fig:macroCorp}}
    \subfigure[Micro-F1 (w/ varying sample size)]{
    \includegraphics[width=0.45\textwidth]{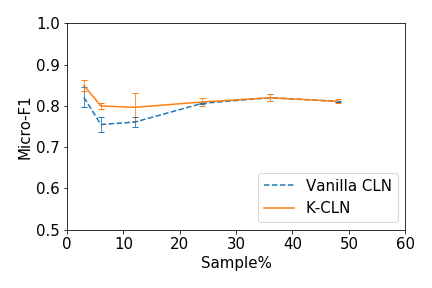}
    \label{fig:microCorpSam}
    }
    \subfigure[Macro-F1 (w/ varying sample size)]{
    \includegraphics[width=0.45\textwidth]{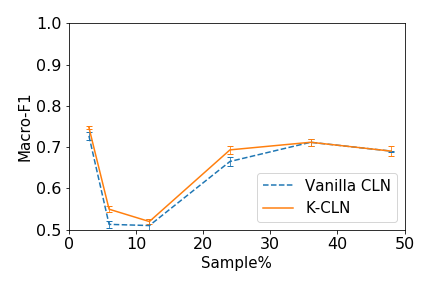}
    \label{fig:macroCorpSam}}
    \caption{\textbf{[Corporate Messages intention prediction (multi-class)]} Learning curves  - \figtop ~ w.r.t. training epochs at 24\% (of total) sample, \figbottom ~ w.r.t. varying sample sizes [best viewed in color].}
\end{minipage}
\end{figure*}

\begin{figure*}
\begin{minipage}{\textwidth}
    \centering
    \subfigure[F1 (w/ epochs)]{
    \includegraphics[width=0.45\textwidth]{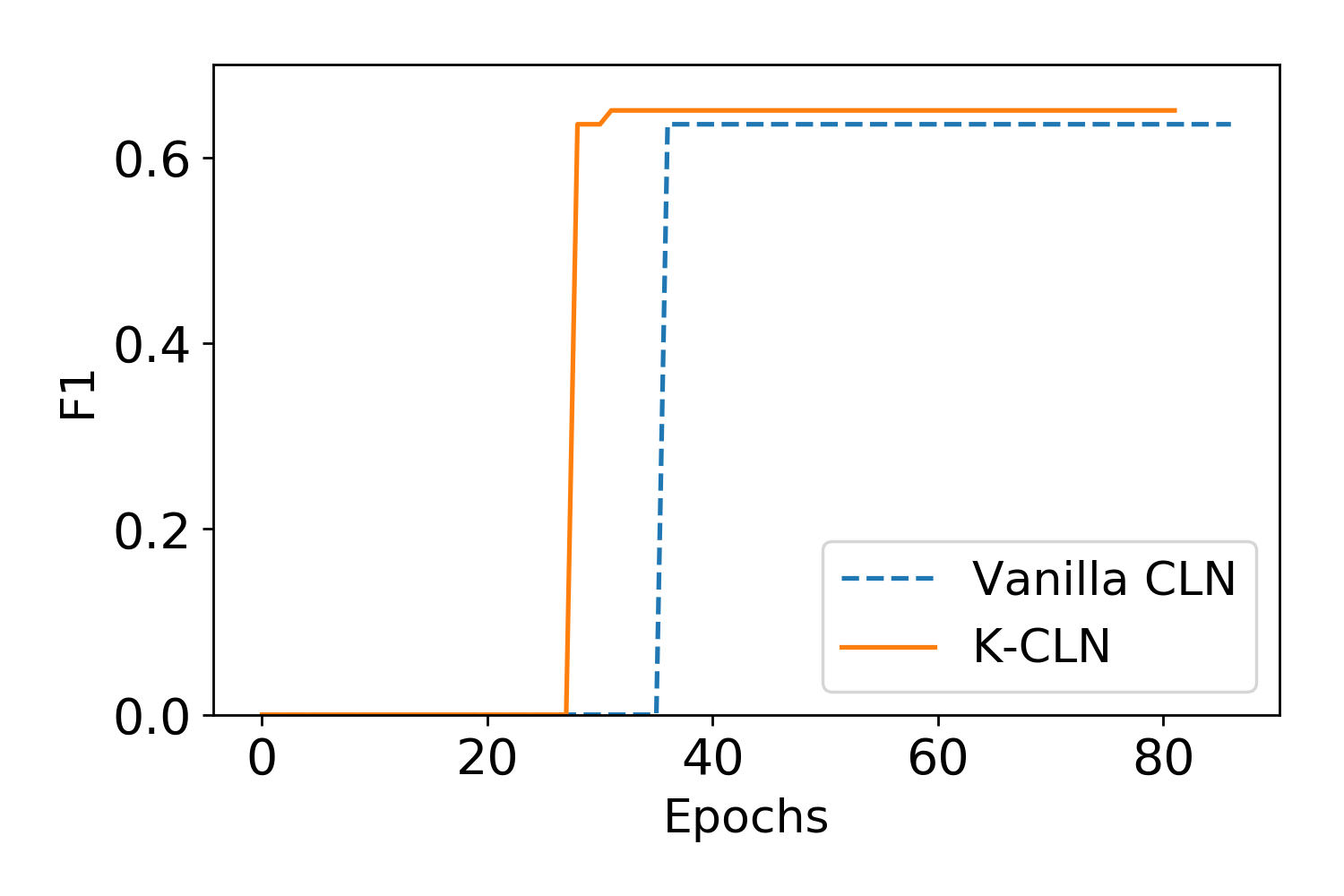}
    \label{fig:debatef1}
    }
    \subfigure[AUC-PR (w/ epochs)]{
    \includegraphics[width=0.45\textwidth]{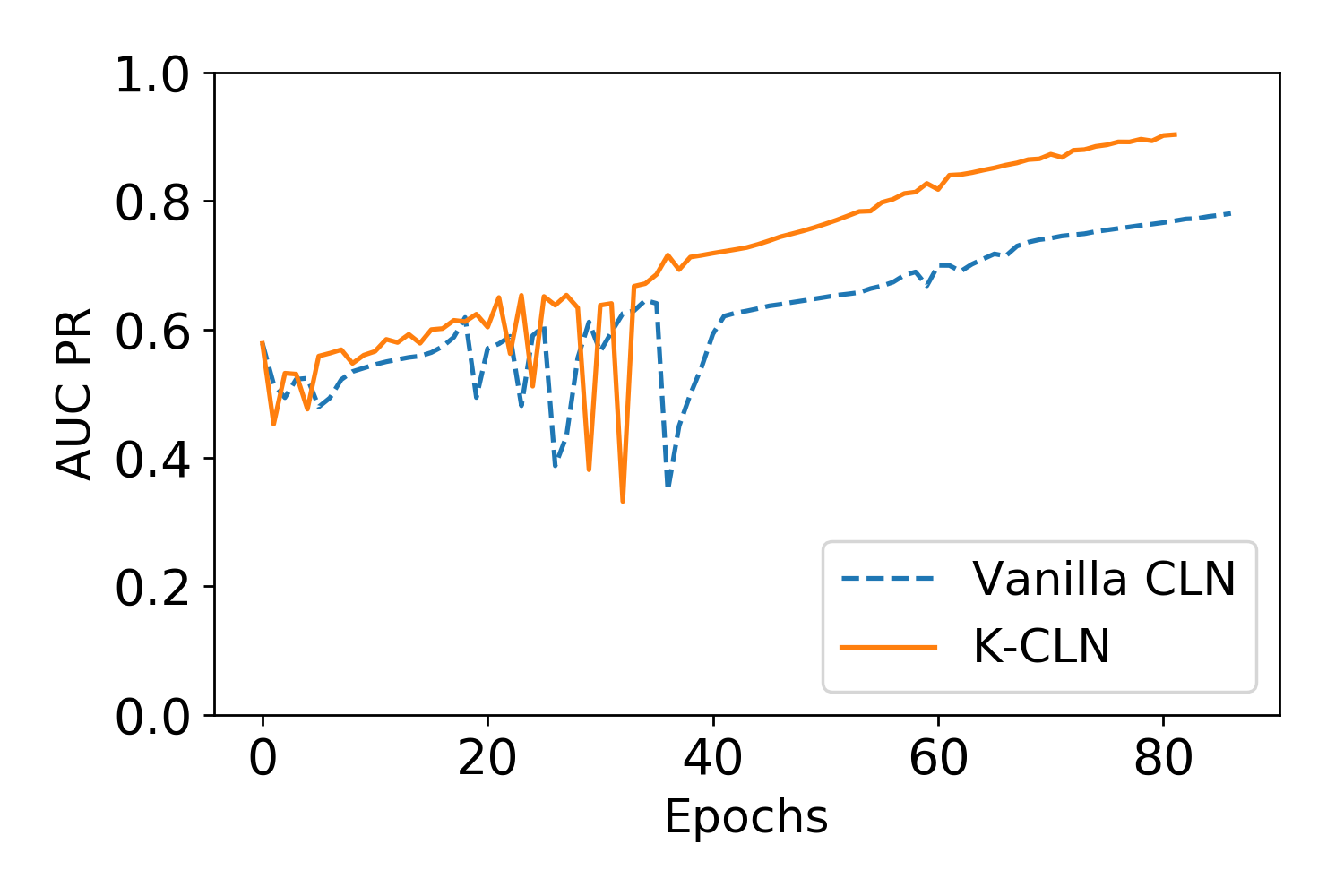}
    \label{fig:debateauc}}
    \subfigure[F1 (w/ varying sample size)]{
    \includegraphics[width=0.45\textwidth]{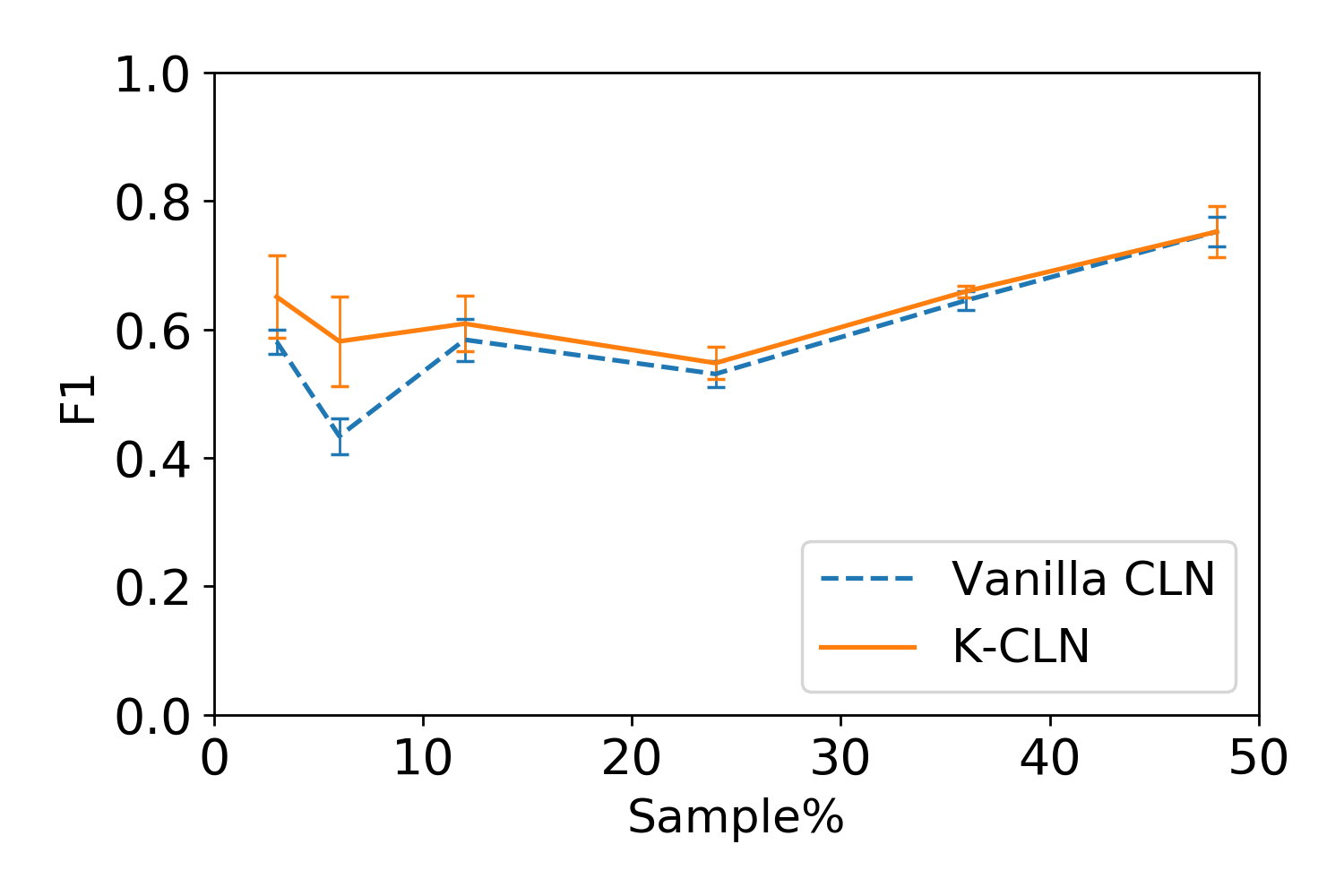}
    \label{fig:debatef1vary}
    }
    \subfigure[AUC-PR (w/ varying sample size)]{
    \centering
    \includegraphics[width=0.45\textwidth]{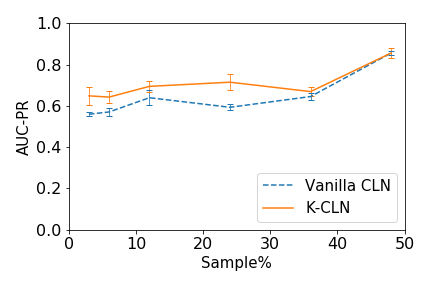}
    \label{fig:debateaucvary}}
    \caption{\textbf{[Internet Social debate stance prediction (binary class)]} Learning curves  - \figtop ~ w.r.t. training epochs at 24\% (of total) sample, \figbottom ~ w.r.t. varying sample sizes [best viewed in color].}
\end{minipage}

\begin{minipage}{\textwidth}
    \centering
    \subfigure[F1 (w/ epochs)]{
    \includegraphics[width=0.45\textwidth]{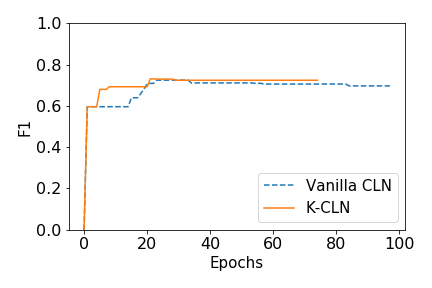}
    \label{fig:f1Social}
    }
    \subfigure[AUC-PR (w/ epochs)]{
    \includegraphics[width=0.45\textwidth]{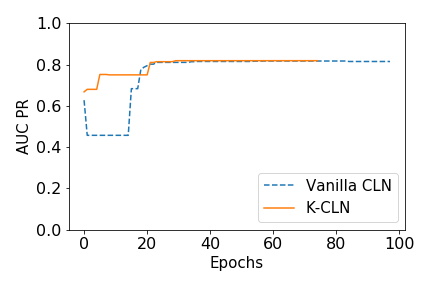}
    \label{fig:aucSocial}}
    \subfigure[F1 (w/ varying samples)]{
    \includegraphics[width=0.45\textwidth]{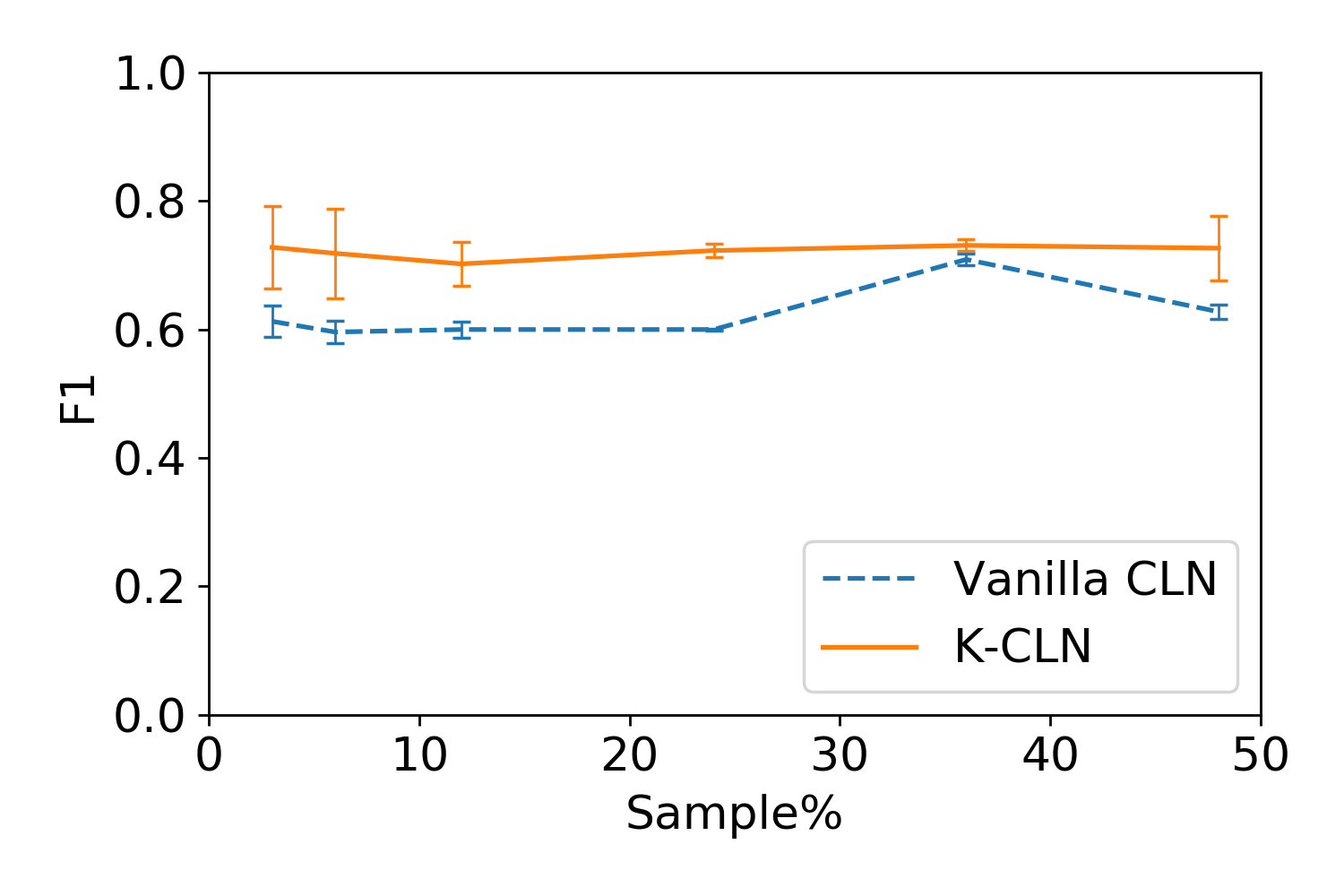}
    \label{fig:f1SocialSam}
    }
    \subfigure[AUC-PR (w/ varying samples)]{
    \includegraphics[width=0.45\textwidth]{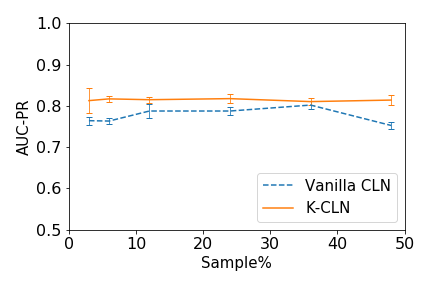}
    \label{fig:aucSocialSam}}
    \caption{\textbf{[Social Network Disaster prediction (binary class)]} Learning curves  - \figtop~ w.r.t. training epochs at 24\% (of total) sample, \figbottom~ w.r.t. varying sample sizes [best viewed in color].}
\end{minipage}
\end{figure*}

\subsection{Experimental Results}
 Recall that our goal is to demonstrate the efficiency and effectiveness of K-CLNs with smaller set of training examples. Hence, we present the aforementioned metrics with varying sample size and with varying epochs and compare our model against \textit{Vanilla CLN}. We split the data sets into a training set and a hold-out test set with 60\%-40\% ratio. For varying epochs we only learn on 40\% of our already split training set (\textit{i.e.}, 24\% of the complete data) to train the model with varying epochs and test on the hold-out test set. Figures \ref{fig:microPub} - \ref{fig:macroPub} illustrate the micro-F1 and the macro-F1 scores  for the \textit{PubMed diabetes} data and Figures \ref{fig:f1Social} - \ref{fig:aucSocial} show the F1 score and AUC-PR for the and social network disaster relevance data. As the figures show, although both K-CLN and Vanilla CLN converge to the same predictive performance, K-CLN converges {\bf significantly faster} (less epochs). For the \textit{corporate messages} and the \textit{internet social debate}, K-CLN not only {\bf converges faster but also has a better predictive performance} than Vanilla CLN as shown in Figures \ref{fig:microCorp} - \ref{fig:macroCorp} and Figures \ref{fig:debatef1} - \ref{fig:debateauc}. The results show that K-CLNs learn more \textit{efficiently} with noisy sparse samples thereby answering \textbf{(Q1)} affirmatively. 

Effectiveness of K-CLN is illustrated by its performance with respect to the varying sample sizes of the training set, especially with low sample size. The intuition is, \textit{domain knowledge should help guide the model to learn better when the amount of training data available is small}. K-CLN is trained on gradually varying sample size from 5\% of the training data (3\% of the complete data) till 80\% of the training data (48\% of complete data) and tested on the hold-out test set. Figures \ref{fig:microPubSam} - \ref{fig:macroPubSam} present the micro-F1 and macro-F1 score for \textit{pubMed diabetes} and Figures \ref{fig:f1SocialSam} - \ref{fig:aucSocialSam} plot the F1 score and AUC-PR for \textit{social network disaster relevance}. It can be seen that K-CLN outperforms Vanilla CLN across all sample sizes, on both metrics, which suggests that the advice is relevant throughout the training phase with varying sample sizes. For \textit{corporate messages}, K-CLN outperforms with small number of samples as shown in the micro-F1 metric (Figure \ref{fig:microCorpSam}) gradually converging to a similar prediction performance with larger samples. Macro-F1 (Figure \ref{fig:macroCorpSam}), however, shows that the performance is similar for both the models across all sample sizes, although K-CLN does perform better with very small samples. Since this is a multi-class classification problem, similar performance in the macro-F1 case suggests that in some classes the advice is not applicable during learning, while it applies well w.r.t. other classes, thereby averaging out the final result.
For \textit{internet social debate stance prediction}, Figures \ref{fig:debatef1vary} - \ref{fig:debateaucvary} present the F1 score and the AUC-PR respectively. K-CLN outperforms the Vanilla CLN on both metrics and thus we can answer \textbf{(Q2)} affirmatively. K-CLNs learn \textit{effectively} with noisy sparse samples.

An obvious question that will arise is -- {\em how robust is our learning system to that of noisy/incorrect advice?} Conversely, {\em how does the choice of $\alpha$ affect the quality of the learned model?}
To answer these questions specifically, we performed an additional experiment on the \textbf{Internet Social Debates} domain by augmenting the learner with incorrect advice. This incorrect advice is essentially created by changing the preferred label of the advice rules to incorrect values (based on our understating). Also, recall that  the contribution of advice is dependent on the trade-off parameter $\alpha$, which controls the robustness of K-CLN to advice quality. Consequently, we experimented with different values of $\alpha$ ($0.2,0.4,\ldots,1.0$), across varying sample sizes.

\begin{figure}[h]
\centering
\subfigure[F1 (varying sample \& $\alpha$)]{
\includegraphics[width = 0.8\textwidth]{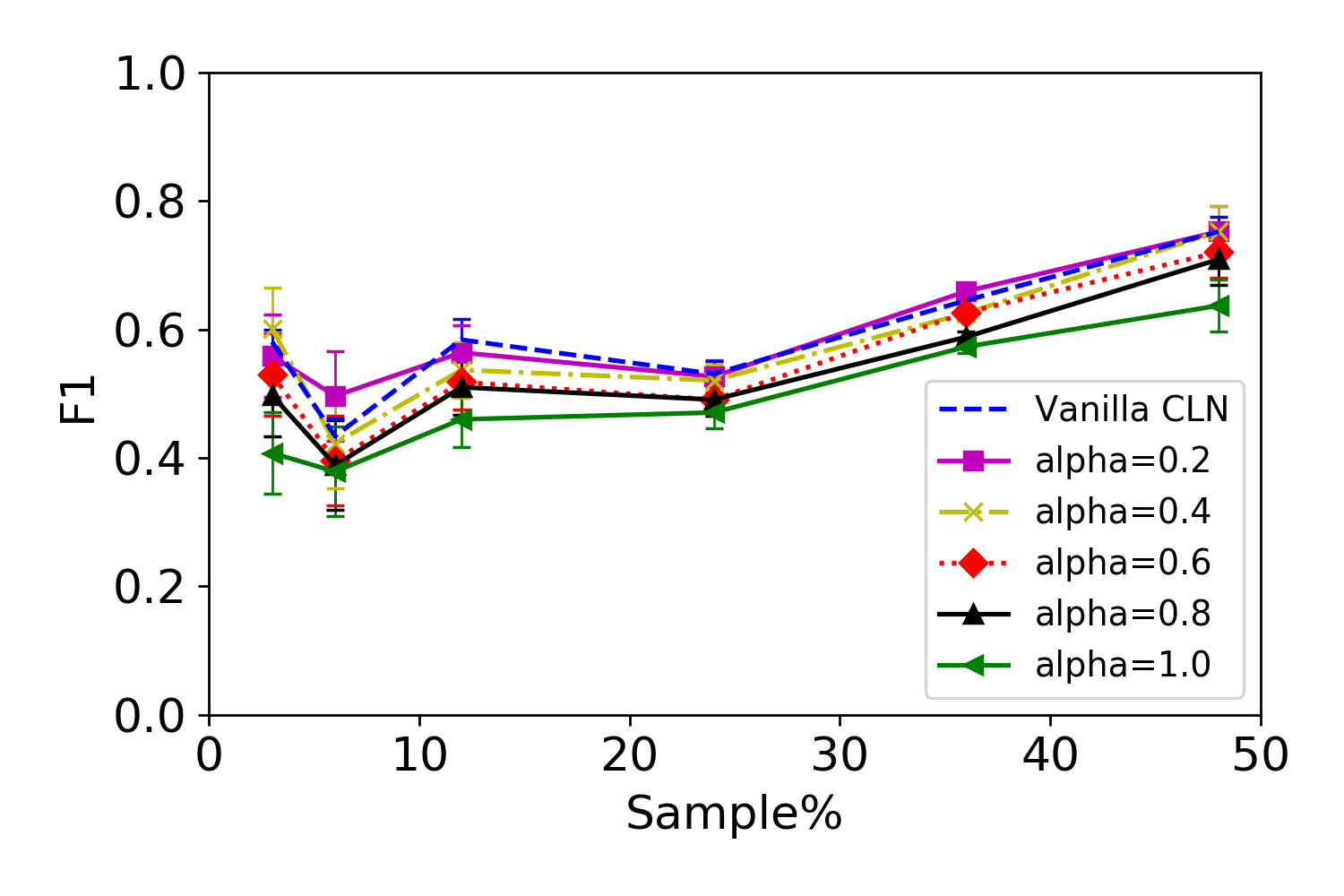}
}
\subfigure[AUC-PR (varying sample \& $\alpha$)]{
\includegraphics[width=0.8\textwidth]{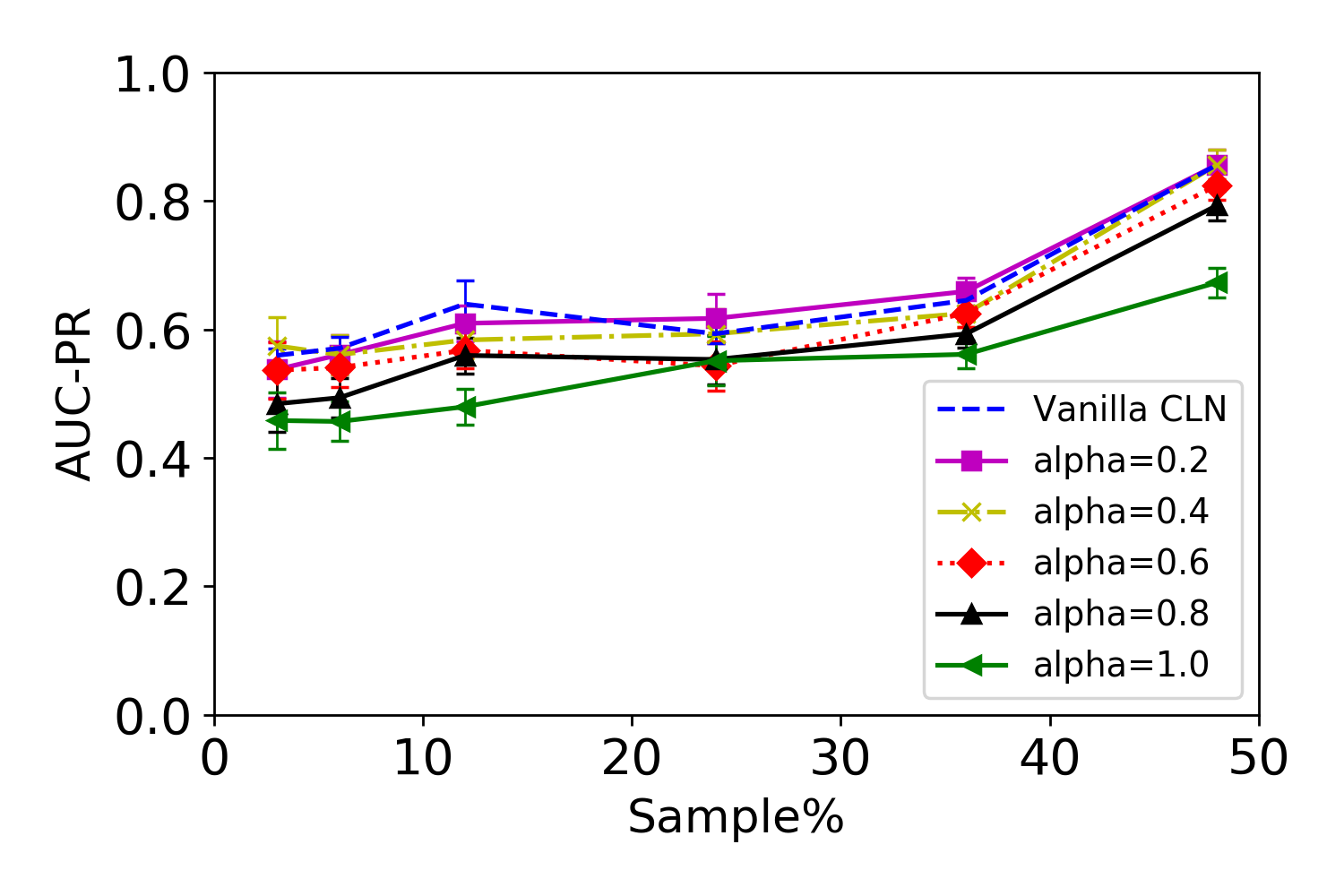}
}
\caption{Performance, F1 and AUC-PR, of K-CLN on \textbf{Internet Social Debates data set} across different sample sizes, with varying \textbf{\textit{trade-off parameter} $\alpha$} (on the advice gradient). Note that the advice here is incorrect/sub-optimal. $\alpha = 0$ has the same performance as no-advice (Vanilla CLN), hence not plotted.}
\label{fig:alphas}
\end{figure}

Figure~\ref{fig:alphas} shows how with higher $\alpha$ values the performance deteriorates due to the effect of noisy advice. $\alpha=0$ is not plotted since the performance is same as no-advice/Vanilla CLN. Note that with reasonably low values of $\alpha = 0.2, 0.4$, the performance does not deteriorate much and is, in fact, better in some samples. Thus with reasonably low values of $\alpha$ K-CLN is robust to quality of advice \textbf{(Q3)}. We picked one domain to present the results of this robustness but have observed similar behavior in all the domains. 

\subsection{Discussion}
It is difficult to quantify correctness or quality of human advice unless, absolute ground truth is accessible in some manner. We evaluate on sparse samples of real data sets with no availability of gold standard labels. Hence, to the best of our capabilities, we have provided the most relevant/useful advice in the experiments aimed at answering (Q1) and (Q2) as indicated in the experimental setup. We emulate noisy advice (for Q3) by flipping/altering the preferred labels of advice rules in the original set of preferences. 

We have shown theoretically, in Proposition~\ref{eq:grad} and \ref{prop:balance}, that the robustness of K-CLN depends on the advice trade-off parameter $\alpha$. We illustrated how it can control the contribution of the data versus the advice towards effective training. We postulate that even in presence of noisy advice, the data (if not noisy) is expected to contribute towards effective learning with a weight of $(1-\alpha)$. Of course, if both the data and advice are noisy the concept is not learnable. Note that this is the case with any learning algorithm where both the knowledge/hypotheses space and the data being incorrect can lead to an incorrect hypothesis.

The experiments \textit{w.r.t.} \textbf{Q3} (Figure~\ref{fig:alphas}) empirically support our theoretical analysis. We found that when $\alpha \leq 0.5$, K-CLN performs well even with noisy advice. In the earlier experiments where we use potentially good advice, we report the results with $\alpha=1$, since the advice gradient is piecewise (affects only a subset of entities/relations). So it is reasonable to assign higher weight to the advice and the contribution of the entities and relations/contexts affected by it, given the advice is noise-free. Also, note that the drop in performance towards very low sample sizes (in Figure~\ref{fig:alphas}) highlights how learning is challenging in the noisy-data and noisy-advice scenario. This aligns with our general understanding of most human-in-the-loop/advice-based approaches in AI. Trade-off between data and advice via a weighted combination of both is a well studied solution in related literature \cite{OdomNatarajan18} and, hence, we adapt the same in our context. Tracking the expertise of humans to infer advice quality is an interesting future research direction.

\section{Conclusion}

We considered the problem of providing guidance for CLNs. Specifically, inspired by treating the domain experts as true domain experts and not CLN experts, we developed a formulation based on {\em preferences}. This formulation allowed for natural specification of guidance. We derived the gradients based on advice and outlined the integration with the original CLN formulation. Our experimental results across different domains clearly demonstrate the effectiveness and efficiency of the approach, specifically in knowledge-rich, data-scarce problems. Exploring other types of advice including feature importances, qualitative constraints, privileged information, etc. is a potential future direction. Scaling our approach to web-scale data is a natural extension. Finally, extending the idea to other deep models remains an interesting direction for future research.



\bibliographystyle{spmpsci}
\bibliography{biblio}

\end{document}